\pdfoutput=1

\documentclass[11pt]{article}

\usepackage{acl}

\usepackage{times}
\usepackage{latexsym}
\usepackage{hyperref}
\usepackage{graphicx}
\usepackage{soul}
\graphicspath{ {figures/} }

\usepackage[colorinlistoftodos]{todonotes}

\usepackage{color} 
\usepackage{ulem}

\usepackage[T1]{fontenc}

\usepackage[utf8]{inputenc}

\usepackage{microtype}

\title{Putting Humans in the Image Captioning Loop}

\author{Aliki Anagnostopoulou$^{1,2}$ \ \ \ \ \ Mareike Hartmann$^1$ \ \ \ \ \ Daniel Sonntag$^{1,2}$ \\
$^1$German Research Center for Artificial Intelligence (DFKI), Germany\\
$^2$Applied Artificial Intelligence (AAI), Oldenburg University, Germany \\
\texttt{\{aliki.anagnostopoulou, mareike.hartmann, daniel.sonntag\}@dfki.de}}

\begin{document}
\maketitle
\begin{abstract}

Image Captioning (IC) models can highly benefit from human feedback in the training process, especially in cases where data is limited. We present work-in-progress on adapting an IC system to integrate human feedback, with the goal to make it easily adaptable to user-specific data. Our approach builds on a base IC model pre-trained on the MS COCO dataset, which generates captions for unseen images. The user will then be able to offer feedback on the image and the generated/predicted caption, which will be augmented to create additional training instances for the adaptation of the model. The additional instances are integrated into the model using step-wise updates, and a sparse memory replay component is used to avoid catastrophic forgetting. We hope that this approach, while leading to improved results, will also result in customizable IC models.

\end{abstract}

\section{Introduction} 

Image Captioning (IC) is the task of generating a natural language description for an image \citep{stefanini-etal}. State-of-the-art IC models are trained in the traditional offline setup, where large amounts of annotated training data are required \citep{zhou2019vlp, li2020oscar, wang2022OFA}. 
This requirement is impractical for models intended to caption user-specific images without large-scale annotations. Here, an \textit{interactive} framework can be used to efficiently adapt a model to new data based on user feedback \cite{ling2017teaching,shen2019learning}. By exploiting  user feedback, models can be trained with less annotated data.
Furthermore, interactivity renders models more user-friendly, 
and the interaction with the user often leads to more trust in the AI/ML-based system \citep{7349687, guo2022trust}.

\begin{figure*}[htp] 
    \centering
    \includegraphics[width=\textwidth]{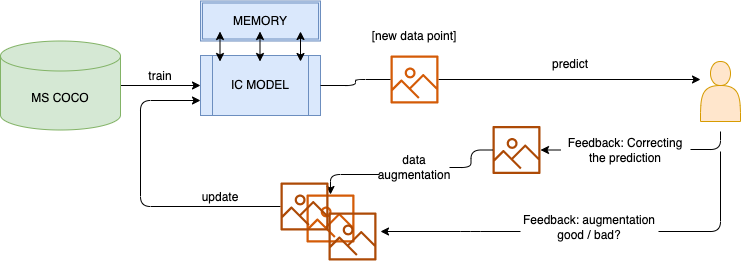}
    \caption{Our proposed pipeline. We pre-train our initial model on MS COCO. The model generates captions for new images, and the user gives feedback on the prediction (by correcting it or marking an area of interest on the image). This feedback is then augmented to create more training instances, and the user can evaluate the quality of the augmentations. The model is then updated accordingly, with a sparse memory replay in order to retain old knowledge.}
    \label{fig:proposal}
\end{figure*}

In the following, we present our work-in-progress on extending an IC model to an interactive setup. Our approach is shown in figure \ref{fig:proposal}. We start with a pretrained IC model (\autoref{main}), which is used to caption new images. The user provides feedback for these captions (\autoref{feedback}), which is then used to generate more training instances via data augmentation (\autoref{aug}). These augmented instances are used to update the model incrementally. In order to retain past knowledge, we employ sparse memory replay (\autoref{cl}). 

In this project, we plan to address four research questions: 

\begin{enumerate}
    \item What type of user feedback is most useful and how can it be collected?
    \item Which data augmentation strategies are most useful to maximize the effect of the user feedback on model performance? 
    \item How helpful is user interaction in the data augmentation process?
    \item How can the feedback best be integrated into the training process?
\end{enumerate}


\section{Experimental setup}

In this section we describe our benchmark strategy, as well as the work-in-progress on data augmentation, model update and evaluation methods, including the human-in-the-loop intersections. The modules described in sections \ref{main} and \ref{aug} are implemented, while the implementation of the ones described in sections \ref{feedback} and \ref{cl} is ongoing.

\subsection{Benchmark strategy}\label{main}

We experiment with a concrete implementation of the interactive approach outlined in \citet{hartmann2022IMLIC}.
As a starting point, we use a PyTorch implementation of the Show, Attend and Tell model \citep{xu2015show}. This architecture consists of a convolutional neural network (CNN) encoder, which is used to extract feature vectors, and a long-short term memory (LSTM) decoder, which generates a caption conditioned on these vectors with attention. The training strategy used is cross-entropy loss. 

In addition, we consider an architecture which requires more supervision, namely the Meshed-Memory (M2) Transformer \citep{cornia2020m2}. 
This model is based on the Transformer architecture proposed by \citet{NIPS2017_3f5ee243}. In the M2 Transformer, the encoder is extended with "slots" for additional, a priori information (\textit{memory}). Additionally, \textit{meshed} cross-attention is performed in the decoder, not only for the last encoding layer, but for all of them. This model requires the additional input of object detections. The model is trained on cross-entropy loss (for pre-training) and reinforcement learning (for fine-tuning).

To pre-train these base models, we use the MS COCO dataset \cite{lin2014microsoft}. 
More specifically, we use the 2014 release, which contains 82,783 training and 40,504 validation images, with five captions per image. We make use of the Karpathy splits \citep{karpathy}.

\subsection{Feedback collection}\label{feedback}
Given the prediction of the base model on a new image, we collect useful feedback from users for both modalities.
This could refer to the correction of a \textit{caption}, but also the drawing of a bounding box around the object on the \textit{image} that was described incorrectly. Additionally, we plan to experiment with feedback collection for the generated augmentations (see \autoref{aug}).

\paragraph{Use case simulation:} Ultimately, we want to apply this method in a real-life scenario where any user can provide feedback to adapt the model to their user-specific data. To simulate this feedback in this early stage, we currently work with an already available dataset, namely VizWiz \citep{vizwiz1, vizwiz2}. VizWiz consists of 23,431 training images, 7,750 validation images and 8,000 test images (39,181 images in total). Each image is annotated with five captions. Since captions for the test set are not publicly available, we test on the validation set and use a small part of the training set as our validation set. 

\subsection{Augmenting the feedback}\label{aug}

\begin{figure*}[htp] 
    \centering
    \includegraphics[width=\textwidth]{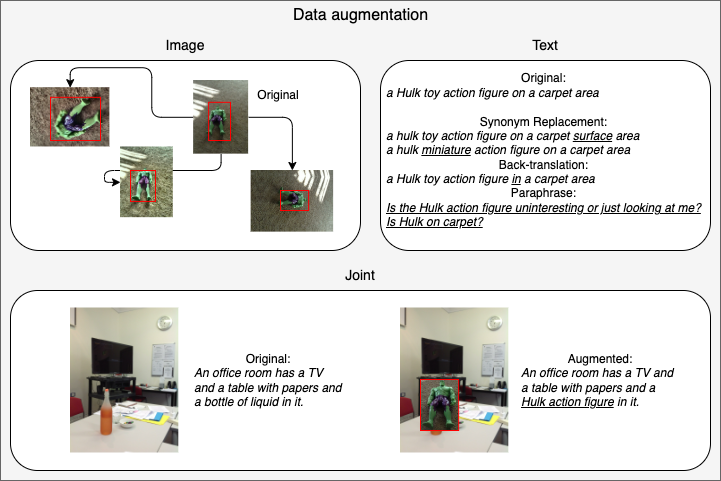}
    \caption{Augmentation examples for image, captions and the joint method. The original data points are from the VizWiz dataset.}
    \label{fig:augmentations}
\end{figure*}

Data augmentation, or synthetic data generation, is a family of techniques that take an initial dataset (often limited in size) and automatically generate more examples \citep{atliha2020text}. Our plan is to augment user feedback in both modalities. Furthermore, we intend to create a novel joint augmentation method, as described below. Some example augmentations from the VizWiz dataset are shown in figure \ref{fig:augmentations}.

\paragraph{Text} 

For the captions, meaning-preserving operations will be used, since our goal is not to introduce noise to the data, but end up with captions that are similar to the feedback provided by the user. We choose three methods:

\begin{enumerate}
    \item Lexical substitution. We follow the EDA \citep{EDA} implementation, which leverages WordNet. 10\% of the tokens of each caption are substituted by a synonym. We generate three augmentations with this method.
    \item Back-translation. We use the ArgosTranslate\footnote{Models are downloaded from: \url{https://www.argosopentech.com/argospm/index/}.} library, translating our English captions to Arabic or Spanish and back to English. As a result, two augmented examples are created.
    \item Paraphrasing with a T5 model \citep{2020t5}, which is specifically fine-tuned for this purpose. With this method, we create five more augmented captions. 
\end{enumerate}

Identical augmentations are discarded. 
The first and third method could provide more augmented examples; we notice, however, that with more examples, the augmentation quality drops significantly, and the number of identical augmentations increases.   
The whole procedure results in about 10 augmentations per caption.

\paragraph{Image}

We use the Albumentations \citep{info11020125} library for image augmentation. We use transformations like rotation, blur, optical distortion, grid distortion, and flip. An additional advantage of the Albumentations library is that it adjusts the bounding boxes with the augmentation. In this manner, feedback provided on the image can be retained.

\paragraph{Joint} We also plan to implement a joint augmentation method, which is based on CutMix \citep{yun2019cutmix} and proposed by \citet{feng-etal-2021-survey}. The idea is to cut objects from different images and insert them in other images. Since this would change the content of the image, the caption should be adjusted accordingly - that is, by addition of a description of the inserted object.

\paragraph{Interaction with users} The examples resulting from the data augmentation step can be used as additional training examples right away. In addition, we consider user interaction with the augmented examples to assure their quality.
More specifically, after user feedback for the prediction is processed and augmented, the user could choose to rank the augmentations (from best to worst), or evaluate them in terms of suitability (\textit{good} / \textit{bad}).

\subsection{Model update and evaluation}\label{cl}

In a real-life application scenario, the user will input images continuously, which means that the system has to be updated multiple times. In cases where a model is trained repeatedly on new data, \textit{catastrophic forgetting} \cite{kirkpatrick2017overcoming} can be observed, namely the degradation of model performance on older tasks when it is trained on new ones. We plan to tackle this problem with a continual or lifelong learning method, and more specifically with a sparse memory replay during training, adapting the idea of \citet{NEURIPS2019_f8d2e80c}: During training, some samples - experiences - are written into the memory. These past experiences are then sparsely replayed while the model is trained on new data.

In order to simulate the step-wise adaptation of the model to new data,
we split the VizWiz dataset in parts of similar size, according to concepts contained in the captions. We follow a naive approach, collecting all noun phrases (NPs) from the captions, and grouping them according to their semantic similarity. We use k-means clustering \citep{Hartigan1979} and pre-trained GloVe word embeddings \citep{pennington2014glove}. All images with captions that contain NPs from the same cluster are then allocated to the same split. 

We plan to train disjointly, namely treating each data split as a new task, which is one of the methods used both by \citet{nguyen2019contcap} (sequential class addition) and \citet{del2020ratt} (disjoint procedure). Evaluation is carried out over individual classes or over all classes, each time the model is trained with a new class/task. 

\paragraph{User evaluation} Apart from evaluating the approach with respect to model performance using automated performance metrics, we plan to evaluate its usefulness and usability for end-users in a human study.

\section{Possible extensions}

Beyond the proposed architecture, we consider some extensions for our pipeline. As mentioned in \autoref{cl}, evaluation from users can potentially point to improvements or the need for the addition of extensions.
While VizWiz, which we use as substitute data in the initial implementation and experimentation stages, constitutes a use case by itself, we do not adapt out implementationto this specific dataset (for example, by employing optical character recognition/detection), as we aim to develop an approach that is applicable to a broad range of user-specific data. Implementing such specific adaptations might further improve performance and can be added on top of our approach depending on the use-case.

Further work can focus on the choice of experiences stored in the memory. This can be done either by integrating the user in the sampling process, or by employing active learning techniques to find the most suitable experiences for future replay.

Last but not least, we can leverage the advantage of interactive systems of learning from fewer labeled instances for cases which annotated data is limited. One such case is multilingual IC. For this reason, an extension of our architecture to support multiple languages looks promising.

\section*{Acknowledgments}

We thank the reviewers for their insightful comments and suggestions. The research was funded by the XAINES project (BMBF, 01IW20005).

\bibliography{main}
\bibliographystyle{acl_natbib}

\end{document}